%% file: camera_ready.tex
\documentclass[sigconf]{acmart}

\AtBeginDocument{%
  \providecommand\BibTeX{{\normalfont B\kern-0.5em{\scshape i\kern-0.25em b}\kern-0.8em\TeX}}}

\usepackage{amsmath,amssymb,amsthm}
\usepackage{booktabs}
\usepackage{graphicx}
\usepackage{tikz}
\usepackage[dvipsnames]{xcolor}

\newtheorem{theorem}{Theorem}
\newtheorem{proposition}[theorem]{Proposition}
\newtheorem{lemma}[theorem]{Lemma}
\newtheorem{corollary}[theorem]{Corollary}
\theoremstyle{definition}
\newtheorem{definition}[theorem]{Definition}
\newtheorem{assumption}{Assumption}
\newtheorem{remark}[theorem]{Remark}

\newcommand{\PA}{\mathrm{PA}}
\newcommand{\DE}{\mathrm{DE}}
\newcommand{\Hcal}{\mathcal{H}}
\newcommand{\Ecal}{\mathcal{E}}
\newcommand{\Xcal}{\mathcal{X}}
\newcommand{\Ycal}{\mathcal{Y}}
\newcommand{\Scal}{\mathcal{S}}
\newcommand{\CME}[2]{C_{#1\mid #2}}
\newcommand{\indep}{\perp\!\!\!\perp}
\newcommand{\HS}{\mathrm{HS}}

\begin{document}

\title{What Environment Invariance Identifies: Minimal Invariant Sets in
Reproducing Kernel Hilbert Spaces}

\author{Ridham Patel}
\affiliation{%
  \department{Computer Science \& Engineering}
  \institution{Indian Institute of Technology Gandhinagar}
  \city{Gandhinagar}
  \state{Gujarat}
  \country{India}
}
\email{ridham.patel@iitgn.ac.in}

\begin{abstract}
Invariant causal prediction identifies the direct causes of a target $Y$ by
searching for covariate sets whose conditional law of $Y$ is stable across
environments. We state that criterion at the level of operators, calling a set
$S$ invariant when the conditional mean embeddings of $Y$ given $X_S$ agree
across environments, and characterise the resulting family exactly: under two
conditions on the environments, which we state separately because they fail
independently, the invariant sets are precisely those containing $\PA(Y)$ and no
descendant of $Y$. Two consequences follow. Invariance does not identify
$\PA(Y)$ uniquely, since every descendant-free superset is invariant as well; and
$\PA(Y)$ is the unique \emph{minimal} invariant set. In finite samples, recovery
requires the non-invariant sets to be separated from invariance by a margin, and
the sample size splits into a screening term logarithmic in the ambient dimension
and a term inherited from conditional mean embedding estimation, which is the
binding one. The characterisation yields two selection rules with different
failure modes, of which the conservative one controls false positives.
Controlled experiments verify each population prediction directly, including
the failure of uniqueness. They also show that invariance testing is a practical
bottleneck: two natural designs are severely miscalibrated, while the selected
test reduces but does not eliminate Type I error inflation.
\end{abstract}


\maketitle

\section{Introduction}
\label{sec:intro}

Causal discovery from observational data becomes difficult when the number of
covariates $d$ is large relative to the sample size $n$. The standard approach
tests conditional independence relations among the
covariates~\cite{spirtes2000causation,zhang2011kernel}, and conditional
independence testing admits no uniformly valid nonparametric
test~\cite{shah2020hardness}. Kernel dependence statistics such as
HSIC~\cite{gretton2005hsic} are consistent in fixed dimension but lose power as
dimension grows~\cite{zhang2023hsic}.

An alternative is to replace independence with invariance. If data are collected
in several environments that differ by interventions, then the mechanism
generating $Y$ from its direct causes is unchanged across environments, while
associations mediated by other variables generally are not. Invariant causal
prediction (ICP)~\cite{peters2016icp} turns this into a procedure: test each
candidate covariate set for stability of the conditional law of $Y$, and report
the intersection of the sets that pass. Invariant risk
minimization~\cite{arjovsky2019irm} pursues the same idea through a learned
representation.

This paper asks two questions about that principle. First, what does invariance
identify? Second, what does it cost to establish that identification from
finite samples? We study both in a formulation where the invariance criterion is
stated at the level of operators.

For a covariate set $S$ and environment $e$, the conditional mean embedding
represents $P_e(Y \mid X_S)$ as a bounded linear operator between reproducing
kernel Hilbert spaces~\cite{song2009cme,park2020measure,muandet2017kernel}.
Calling $S$ invariant when these operators agree across environments gives a
criterion needing no parametric model for $P_e(Y\mid X_S)$ and no choice of
predictor, and expressing departure from invariance as a single
Hilbert--Schmidt norm (Section~\ref{sec:criterion}).

Our first result is negative and constrains what any invariance-based method can
achieve. Under the standard assumption that no environment intervenes on the
mechanism of $Y$, the parent set $\PA(Y)$ is invariant. It is not, however, the
only invariant set. If $Z$ is a non-descendant of $Y$, the local Markov
property~\cite{pearl2009causality,peters2017elements} gives $P(Y \mid \PA(Y), Z) = P(Y \mid \PA(Y))$, so $\PA(Y) \cup \{Z\}$ is
invariant as well (Proposition~\ref{prop:superset}). Uniqueness therefore fails
for reasons internal to the causal semantics rather than through any deficiency
of an estimator. What does hold is that $\PA(Y)$ is the unique minimal invariant
set (Theorem~\ref{thm:minimal}), and this requires conditions on the
environments.

Those conditions are often bundled into a single hypothesis that environments are
``sufficiently diverse.'' Separating them is useful because they do different
work and fail independently. \emph{Descendant activation} excludes sets that
contain descendants of $Y$; \emph{parent activation} excludes strict subsets of
$\PA(Y)$. Section~\ref{sec:experiments} exhibits each failing on its own, with
the consequence the analysis predicts in each case.

An identifiability theorem quantifies over population objects, and says nothing
about finite samples unless those objects can be estimated. Conditional mean
embeddings are estimable, at rates governed by a source condition and an
effective dimension, and these rates are in general slower than
$n^{-1/2}$~\cite{li2022optimal,li2024towards}. We therefore state sample
complexity in a form that separates the two dependencies: $d$ enters
logarithmically through a sparse screening step, while the dependence on $n$ is
inherited from embedding estimation. Recovery additionally requires the
non-invariant sets to be quantitatively separated from invariance, which we state
as a margin condition rather than leave implicit.

The analysis makes numerical predictions, and a controlled experiment run where
estimation error is small must reproduce them. At $n=1000$ with $d=8$ every
per-claim prediction is reproduced at rate $1.00$, including the prediction that
supersets of $\PA(Y)$ are accepted (Section~\ref{sec:exp-verify}). A separate
calibration study shows why this population agreement does not by itself imply a
reliable finite-sample procedure: two natural test designs have Type I errors of
$0.493$ and $1.000$ against a nominal $0.05$, and even the selected design remains
approximately rather than exactly calibrated (Section~\ref{sec:test-design}).

Because the operator criterion turns out to be equivalent to the conditional
independence ICP already tests (Proposition~\ref{prop:ci}), it is worth saying
precisely what is added here rather than leaving it to be inferred. ICP
establishes a conservative coverage guarantee: the intersection of the accepted
sets is contained in $\PA(Y)$. We instead characterise the entire invariant
family, showing in Theorem~\ref{thm:minimal} that it consists exactly of the
sets containing $\PA(Y)$ and no descendant of $Y$. That characterisation is what
makes the failure of uniqueness visible, identifies $\PA(Y)$ as the family's
unique minimal element rather than its only member, and separates the two
environment conditions responsible for the two halves of the description. It
also changes what a finite-sample analysis must deliver: not a bound on excess
risk, but a margin separating non-invariant sets from invariance, together with
an estimation error small enough to resolve it.

In summary, Section~\ref{sec:criterion} gives an operator formulation of the
invariance criterion and its equivalence to $Y \indep E \mid X_S$.
Section~\ref{sec:identifiability} characterises the invariant family exactly, as
$\Scal_{\mathrm{inv}} = \{S : \PA(Y)\subseteq S,\ S\cap\DE(Y)=\emptyset\}$
under two separately stated environment conditions, from which it follows that
invariant sets are not unique and that $\PA(Y)$ is the unique minimal one.
Section~\ref{sec:complexity} gives a finite-sample recovery guarantee under an
explicit margin condition, with a sample complexity that separates a $\log d$
screening term from the embedding-estimation term that binds. Finally,
Sections~\ref{sec:algorithm} and~\ref{sec:experiments} test each claim of the
analysis directly and examine the calibration of the invariance test.

\section{Related Work}
\label{sec:related}

ICP~\cite{peters2016icp} tests whether the regression of $Y$ on $X_S$ is stable
across environments and returns the intersection of the accepted sets, which is
contained in $\PA(Y)$ with coverage guarantees; extensions treat nonlinear
models~\cite{heinzedeml2018nonlinearicp}, sequential
data~\cite{pfister2019seqicp}, and transfer
learning~\cite{rojascarulla2018invariant}, and related formulations include
joint causal inference~\cite{mooij2020jci} and anchor
regression~\cite{rothenhausler2021anchor}. The criterion we study is a
nonparametric instance of ICP's, made exact by Proposition~\ref{prop:ci}.

IRM~\cite{arjovsky2019irm} seeks a representation admitting a single optimal
predictor across environments, through a gradient-norm surrogate whose behaviour
has been examined closely: the surrogate need not capture the intended
invariance~\cite{kamath2021does}, the objective degrades outside its assumed
regime~\cite{rosenfeld2021risks}, and gains over empirical risk minimization are
often not reproducible~\cite{gulrajani2021search}.

Conditional mean embeddings represent conditional distributions as
operators~\cite{song2009cme,grunewalder2012cme}. Measure-theoretic foundations
are given in~\cite{park2020measure} and a survey
in~\cite{muandet2017kernel}; characteristic kernels determine when the
embeddings are
injective~\cite{sriperumbudur2010hilbert,sriperumbudur2011universality}. Their
estimation theory~\cite{li2022optimal,li2024towards} gives rates in terms of a
source condition and an effective dimension, and those rates enter our
finite-sample analysis directly.

Multiple kernel learning under an $\ell_1$ constraint admits complexity bounds
that scale logarithmically in the number of base
kernels~\cite{cortes2010generalization,kloft2011local,kloft2012convergence}, and
for additive kernels coincides with the group
lasso~\cite{bach2008consistency,yuan2006model}. We use it for screening, where
logarithmic dependence on $d$ is what the regime $d \gg n$ requires.

\section{Preliminaries}
\label{sec:prelim}

\subsection{Structural causal models and environments}

Let $X = (X_1,\dots,X_d)$ and $Y$ be generated by a structural causal model over
a directed acyclic graph $G$,
\[
X_i := f_i(\PA(X_i), U_i), \qquad Y := f_Y(\PA(Y), U_Y),
\]
with jointly independent exogenous noise
variables~\cite{pearl2009causality,peters2017elements}. We write $\PA(Y)$ for the
set of direct causes of $Y$, $\DE(Y)$ for its descendants, and $X_S$ for the
subvector indexed by $S \subseteq \{1,\dots,d\}$.

Data arrive from a finite set of environments $\Ecal$, and we write $E$ for the
random variable indexing the environment, with a fixed distribution placing
positive mass on each $e \in \Ecal$. Each environment $e$ induces a joint law
$P_e(X,Y)$ obtained by intervening on a subset of the structural equations. The
following assumption is what makes invariance informative about causal structure.

\begin{assumption}[Modularity]
\label{ass:modularity}
No environment intervenes on the structural equation of $Y$: the function $f_Y$
and the law of $U_Y$ are the same in every $e \in \Ecal$.
\end{assumption}

Assumption~\ref{ass:modularity} is the standard autonomy
postulate~\cite{pearl2009causality,peters2017elements}. It permits interventions
on the parents of $Y$, which change the marginal law of $X_{\PA(Y)}$ while
leaving $P(Y \mid X_{\PA(Y)})$ unchanged. That distinction matters in
Section~\ref{sec:identifiability}, where interventions on parents are what
distinguishes $\PA(Y)$ from its subsets.

\subsection{Conditional mean embeddings}

The invariance criterion must compare the laws $P_e(Y \mid X_S)$ across
environments, so it needs a representation of them that presupposes no parametric
model. Conditional mean embeddings provide one.

Let $k_\Xcal$ and $k_\Ycal$ be bounded measurable positive definite kernels on
$\Xcal$ and $\Ycal$, with reproducing kernel Hilbert spaces $\Hcal_\Xcal$ and
$\Hcal_\Ycal$; we assume $k_\Xcal(x,x)\le 1$ and $k_\Ycal(y,y)\le 1$ throughout.
A kernel is \emph{characteristic} if the mean embedding
$P \mapsto \mathbb{E}_{Z \sim P}[k(Z,\cdot)]$ is injective on Borel probability
measures~\cite{sriperumbudur2010hilbert}; the Gaussian and Laplace kernels on
$\mathbb{R}^m$ are
characteristic~\cite{sriperumbudur2010hilbert,sriperumbudur2011universality}. Injectivity is what allows equality of
embeddings to be read back as equality of distributions, and it is used in
Proposition~\ref{prop:ci}.

The conditional mean embedding of $Y$ given $X_S$ is the operator
$\CME{Y}{X_S} : \Hcal_{\Xcal_S} \to \Hcal_\Ycal$ satisfying
\begin{equation}
\label{eq:cme}
\mathbb{E}[g(Y) \mid X_S = x]
= \langle g, \CME{Y}{X_S} k_{\Xcal_S}(x,\cdot)\rangle_{\Hcal_\Ycal}
\quad \text{for all } g \in \Hcal_\Ycal ,
\end{equation}
when such an operator exists~\cite{song2009cme,park2020measure}. Equation
\eqref{eq:cme} says that applying the operator to the feature map of $x$ returns
the embedding of the conditional distribution at $x$. We use the construction
of~\cite{park2020measure}, under which \eqref{eq:cme} holds $P_{X_S}$-almost
surely without requiring $\mathbb{E}[g(Y)\mid X_S=\cdot\,] \in \Hcal_{\Xcal_S}$.

\begin{remark}[Domain and codomain]
\label{rem:typing}
$\CME{Y}{X_S}$ maps $\Hcal_{\Xcal_S}$ into $\Hcal_\Ycal$, and these are distinct
spaces. Expressions of the form $I - \CME{Y}{X_S}$, with $I$ the identity on
$\Hcal_{\Xcal_S}$, are therefore not well defined. Any measure of departure from
invariance must be built from differences of operators that share a domain and a
codomain, which is how Definition~\ref{def:discrepancy} is constructed.
\end{remark}

\section{The Invariance Criterion}
\label{sec:criterion}

\subsection{Definition}

The criterion compares the conditional law of $Y$ given $X_S$ across
environments. In the operator representation this is an equality of operators.

\begin{definition}[Invariant set]
\label{def:invariance}
A set $S \subseteq \{1,\dots,d\}$ is \emph{invariant} if
\[
\CME{Y}{X_S}^{(e)} = \CME{Y}{X_S}^{(e')} \qquad \text{for all } e,e' \in \Ecal .
\]
We write $\Scal_{\mathrm{inv}}$ for the family of invariant sets.
\end{definition}

To quantify how far a set is from invariance we need a scalar. Averaging the
operators across environments and measuring the spread gives one, and
Remark~\ref{rem:typing} dictates its form.

\begin{definition}[Invariance discrepancy]
\label{def:discrepancy}
Let $\bar{C}_S := \mathbb{E}_e[\CME{Y}{X_S}^{(e)}]$ and let
\[
Z_S := \big(\CME{Y}{X_S}^{(e)} - \bar{C}_S\big)_{e \in \Ecal} :
\Hcal_{\Xcal_S} \to \textstyle\bigoplus_{e \in \Ecal} \Hcal_\Ycal .
\]
The invariance discrepancy is
$D(S) := \|Z_S\|_{\HS}^2 = \sum_{e} \|\CME{Y}{X_S}^{(e)} - \bar{C}_S\|_{\HS}^2$.
\end{definition}

Each summand maps $\Hcal_{\Xcal_S}$ to $\Hcal_\Ycal$, so $Z_S$ is well defined,
and $S \in \Scal_{\mathrm{inv}}$ if and only if $D(S) = 0$. The quantity $D(S)$
is the population object that the test of Section~\ref{sec:test-design} estimates
and that the margin condition of Section~\ref{sec:complexity} bounds away from
zero.

\subsection{Relation to conditional independence}

Before analysing what Definition~\ref{def:invariance} identifies, we establish
what it is equivalent to. This determines whether the operator formulation
introduces a new identification principle or restates an existing one.

\begin{proposition}[Equivalence]
\label{prop:ci}
Suppose $k_\Ycal$ is characteristic, the embeddings $\CME{Y}{X_S}^{(e)}$ exist
for every $e$, and the laws $P_e(X_S)$ are mutually absolutely continuous with
common support. Then $S$ is invariant in the sense of
Definition~\ref{def:invariance} if and only if $Y \indep E \mid X_S$.
\end{proposition}

The proof is in Appendix~\ref{app:criterion}. The conditional independence
$Y \indep E \mid X_S$ is the criterion ICP tests, so the operator formulation
restates that criterion rather than strengthening it. Its advantages are
representational: it requires no model for $P_e(Y\mid X_S)$, and it supplies the
scalar $D(S)$.

The hypotheses of Proposition~\ref{prop:ci} are not technical decoration, and two
of them constrain how the criterion may be used.

\begin{remark}[The kernel on $Y$ determines what is compared]
\label{rem:kernel-conditions}
If $k_\Ycal$ is not characteristic the equivalence fails. The kernel
$k_\Ycal \equiv 0$ is bounded and positive definite, and makes every set
invariant. A linear kernel on $Y$ makes Definition~\ref{def:invariance}
equivalent to invariance of $\mathbb{E}[Y \mid X_S]$ alone, which is the
first-moment condition underlying IRM~\cite{arjovsky2019irm} and is strictly
weaker than $Y \indep E \mid X_S$.
\end{remark}

\begin{remark}[Common support]
\label{rem:support}
Interventions shift $P_e(X_S)$ and can therefore separate the supports of
different environments. Where one environment places no mass, its conditional law
is unconstrained, and invariance is meaningful only on the intersection of the
supports. This is a property of the estimand, and Section~\ref{sec:test-design}
shows that it also governs which estimators of $D(S)$ remain calibrated.
\end{remark}

The criterion is stated for a covariate set $S$. Formulations such as IRM instead
constrain a learned representation $\Phi$, and it has been asked whether such
constraints are trivially satisfiable by reparameterisation.
Proposition~\ref{prop:collapse} in Appendix~\ref{app:criterion} records that the
criterion depends on $\Phi$ only through $\sigma(\Phi(X))$, so a bijective
reparameterisation leaves it unchanged and a constant representation satisfies it
exactly when the marginal law of $Y$ is environment-invariant. The question does
not arise below, where identifiability is stated over covariate subsets.

\section{What Invariance Identifies}
\label{sec:identifiability}

\subsection{Invariance does not determine \texorpdfstring{$\PA(Y)$}{PA(Y)} uniquely}

We begin with the positive statement, which follows from
Assumption~\ref{ass:modularity} alone.

\begin{proposition}[Invariance of $\PA(Y)$]
\label{prop:pa-invariant}
Under Assumption~\ref{ass:modularity}, $\PA(Y) \in \Scal_{\mathrm{inv}}$.
\end{proposition}

The parent set is therefore always among the invariant sets. It is not the only
one, and the reason is the local Markov property rather than any property of an
estimator.

\begin{proposition}[Supersets are invariant]
\label{prop:superset}
Under Assumption~\ref{ass:modularity}, let $S$ satisfy $\PA(Y) \subseteq S$ and
$S \cap \DE(Y) = \emptyset$. Then $S \in \Scal_{\mathrm{inv}}$.
\end{proposition}

\begin{proof}[Proof sketch]
Write $R := S \setminus \PA(Y)$. The local Markov
property~\cite{pearl2009causality,peters2017elements} gives
$Y \indep X_R \mid X_{\PA(Y)}$ in every environment, so that
$P_e(Y \mid X_S) = P_e(Y \mid X_{\PA(Y)})$, whose right-hand side is
environment-invariant by Proposition~\ref{prop:pa-invariant}. The full argument
is in Appendix~\ref{app:identifiability}.
\end{proof}

\begin{figure}[t]
\centering
\input{figures/lattice}
\caption{The invariant family of Theorem~\ref{thm:minimal}, ordered by
inclusion. Every superset avoiding descendants of $Y$ is invariant
(Proposition~\ref{prop:superset}), so the family is upward-closed rather than a
single point, and $\PA(Y)$ is distinguished only by being its smallest member.
Assumption~\ref{ass:parent} excludes the sets below $\PA(Y)$;
Assumption~\ref{ass:descendant} excludes the branch containing a descendant $D$.}
\Description{A set-inclusion diagram with the parent set at its minimal node,
invariant non-descendant supersets above it, and excluded subsets and
descendant-containing sets on separate branches.}
\label{fig:lattice}
\end{figure}
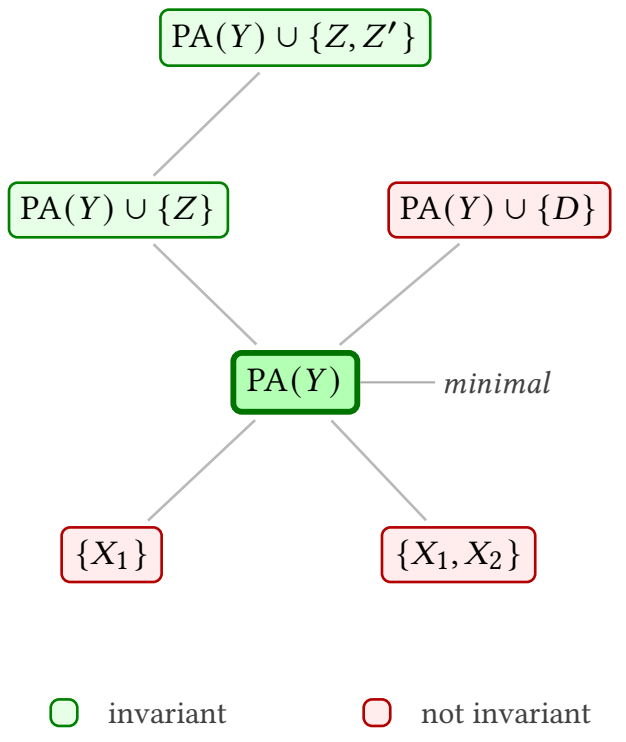

Proposition~\ref{prop:superset} rules out any claim that $\PA(Y)$ is the unique
invariant set. Whenever a non-descendant of $Y$ is available, which is the generic
situation, the invariant family contains a chain of sets strictly above $\PA(Y)$.
Any procedure whose only criterion is invariance will accept all of them.
Figure~\ref{fig:lattice} shows the resulting structure, and
Section~\ref{sec:exp-verify} measures the rate at which these supersets are
accepted.

One guarantee survives without further conditions.

\begin{corollary}[Conservative guarantee]
\label{cor:icp}
Under Assumption~\ref{ass:modularity},
$\bigcap_{S \in \Scal_{\mathrm{inv}}} S \subseteq \PA(Y)$.
\end{corollary}

This is immediate, since $\PA(Y)$ belongs to the intersected family, and it is
the guarantee ICP provides~\cite{peters2016icp}: the reported set contains no
false positives, and may be empty.

\subsection{Minimality, under two conditions}

A sharper conclusion requires conditions on the environments. Two are needed, and
they do different work: one excludes sets containing descendants of $Y$, the
other excludes strict subsets of $\PA(Y)$. We state them separately because they
are violated by different experimental designs. These assumptions do not derive
environment diversity from weaker graphical conditions. Instead, they state the
two population distinctions that environments must reveal, allowing the theorem
to identify which distinction is responsible when minimality fails.

\begin{assumption}[Descendant activation]
\label{ass:descendant}
For every $S$ with $S \cap \DE(Y) \neq \emptyset$ there exist $e,e' \in \Ecal$
with $P_e(Y \mid X_S) \neq P_{e'}(Y \mid X_S)$. It is sufficient that for each
$D \in \DE(Y)$, some environment intervenes on a mechanism lying on a directed
path from $Y$ to $D$.
\end{assumption}

\begin{assumption}[Parent activation]
\label{ass:parent}
For every $S \not\supseteq \PA(Y)$ there exist $e,e' \in \Ecal$ with
$P_e(Y \mid X_S) \neq P_{e'}(Y \mid X_S)$. It is sufficient that for each
$j \in \PA(Y)$ some environment shifts
$P(X_j \mid X_{\PA(Y) \setminus \{j\}})$, together with a non-cancellation
condition ensuring that the shift is not annihilated when $X_j$ is marginalised
out of $f_Y$.
\end{assumption}

The non-cancellation clause in Assumption~\ref{ass:parent} is a faithfulness-type
condition of the kind ICP also requires, and it is not implied by the graph. We
state it explicitly because it can fail in ordinary models: if $f_Y$ saturates,
a shift in one parent can push $f_Y$ into a region where shifts in the remaining
parents no longer affect $Y$. Section~\ref{sec:experiments} exhibits this failure
and its consequence.

\begin{theorem}[$\PA(Y)$ is the unique minimal invariant set]
\label{thm:minimal}
Under Assumptions~\ref{ass:modularity}, \ref{ass:descendant}
and~\ref{ass:parent},
\begin{equation}
\label{eq:inv-family}
\Scal_{\mathrm{inv}} = \{\, S : \PA(Y) \subseteq S,\; S \cap \DE(Y) = \emptyset \,\}.
\end{equation}
Consequently $\PA(Y)$ is the unique minimal element of $\Scal_{\mathrm{inv}}$
under inclusion, and $\bigcap_{S \in \Scal_{\mathrm{inv}}} S = \PA(Y)$.
\end{theorem}

Equation~\eqref{eq:inv-family} describes the invariant family completely, and
each assumption contributes one of its two defining conditions:
Assumption~\ref{ass:parent} gives $\PA(Y) \subseteq S$ and
Assumption~\ref{ass:descendant} gives $S \cap \DE(Y) = \emptyset$. Dropping
either one enlarges the family in a specific and observable way, which is what
makes the two conditions separately testable in Section~\ref{sec:experiments}.

Minimality can equivalently be stated as a fixed-point property, which is the
form the algorithm of Section~\ref{sec:algorithm} uses:
$\PA(Y)$ is the unique set in $\Scal_{\mathrm{inv}}$ from which no element can be
deleted while preserving invariance. Appendix~\ref{app:identifiability} gives
this formulation, and records why invariance must be treated as a property of a
set rather than of individual covariates.

\section{Estimation and Sample Complexity}
\label{sec:complexity}

Section~\ref{sec:identifiability} concerns population operators. We now ask what
is achievable when those operators are estimated, since this determines the
sample complexity of recovery.

\subsection{Estimating the operators}

Given $n_e$ samples from environment $e$, the regularised empirical conditional
mean embedding~\cite{song2009cme,grunewalder2012cme} is
\begin{equation}
\label{eq:cme-est}
\widehat{C}^{(e)}_{Y \mid X_S}
= \Phi_Y^{(e)} \big(K^{(e)} + n_e\lambda I\big)^{-1}\Psi^{(e)\top},
\end{equation}
where $K^{(e)}$ is the Gram matrix of $X_S$ in environment $e$. The accuracy of
\eqref{eq:cme-est} is governed by a source condition and an effective-dimension
parameter~\cite{li2022optimal,li2024towards}. Since the sample complexity below
depends on this accuracy and on nothing else about the estimator, we state what
we assume of it rather than leave it implicit.

\begin{assumption}[Uniform embedding error]
\label{ass:cme}
Fix a candidate pool $\Scal_{\mathrm{pool}}$ of size $K$. There are constants
$\beta \in (0,1]$ and $c \in (0,1]$ such that, for every $S \in
\Scal_{\mathrm{pool}}$ and $e \in \Ecal$, the target $\CME{Y}{X_S}^{(e)}$ lies
in the $\beta$-power range of the covariance operator of $X_S$ (the source
condition) and that operator has eigenvalues decaying at rate $i^{-1/c}$ (the
effective-dimension condition). Both conditions are the standard pair for
regularised least squares in an RKHS~\cite{caponnetto2007optimal}, transferred
to conditional mean embeddings
by~\cite{li2022optimal,li2024towards}. With $\lambda_n \asymp n^{-1/(\beta+c)}$ and
$n := \min_e n_e$, there is $\varepsilon_n$ with
\[
\Pr\Big[\;\sup_{S \in \Scal_{\mathrm{pool}},\, e \in \Ecal}
\big\|\widehat{C}^{(e)}_{Y\mid X_S} - \CME{Y}{X_S}^{(e)}\big\|_{\HS}
\le \varepsilon_n \Big] \ge 1-\delta_2,
\]
where $\varepsilon_n \asymp \sqrt{\log(KE/\delta_2)}\; n^{-b}$ and
$b = \beta/(2(\beta+c)) \le 1/2$.
\end{assumption}

Three features matter below. First, the supremum is taken over the screened
pool, not over all $2^d$ subsets, so $K$ rather than $d$ enters the union
bound. Second, the sample size is that of the smallest environment, since every
operator must be estimated well. Third, $b \le 1/2$, with equality only in the
most favourable regime, so the error is in general slower than parametric. We do not prove
Assumption~\ref{ass:cme}: it is the form in which the rates
of~\cite{li2022optimal,li2024towards} enter, and what follows is conditional on
it.

\subsection{Complexity of the hypothesis class}

We record the complexity of an RKHS ball, both because it is used for the
screening stage and because its correct form is easy to misstate.

The relevant notion is the Rademacher complexity of the hypothesis
class~\cite{bartlett2002rademacher}, which for an RKHS ball takes the following
form.

\begin{lemma}[Rademacher complexity of an RKHS ball]
\label{lem:rademacher}
Let $\Hcal$ have kernel $k$ with $k(x,x) \le 1$, and let $B_\Hcal$ be its unit
ball. Then
\[
\widehat{\mathfrak{R}}_n(B_\Hcal)
= \frac{1}{n}\,\mathbb{E}_\sigma\Big\|\sum_{i=1}^n \sigma_i k(x_i,\cdot)\Big\|_\Hcal
\;\le\; \frac{\sqrt{\operatorname{Tr}K}}{n},
\]
and $\widehat{\mathfrak{R}}_n(B_\Hcal) \ge \sqrt{\operatorname{Tr}K/2}\,/\,n$ by
the Khintchine--Kahane inequality~\cite{latala1994khinchin}.
\end{lemma}

Since $\operatorname{Tr}K \le n$ under $k(x,x)\le 1$, Lemma~\ref{lem:rademacher}
gives $\widehat{\mathfrak{R}}_n(B_\Hcal) \le n^{-1/2}$. The matching lower bound
shows the quantity cannot be improved by a sharper argument; it also shows why
this complexity is not the quantity that controls support recovery, a point we
return to in Section~\ref{sec:margin}.

For screening we use $\ell_1$-constrained multiple kernel learning over $d$ base
kernels, one per covariate, whose complexity scales logarithmically in the number
of kernels~\cite{cortes2010generalization,kloft2011local,kloft2012convergence}.
For the class $\Hcal_{s,d}$ of functions supported on at most $s$ of the $d$
blocks~\cite{cortes2010generalization,bach2008consistency,yuan2006model},
\begin{equation}
\label{eq:mkl-bound}
\mathfrak{R}_n(\Hcal_{s,d}) \;\lesssim\; \sqrt{\frac{s\log d}{n}} .
\end{equation}
This is the only place where the ambient dimension enters, and it enters
logarithmically.

\subsection{Recovery requires a margin}
\label{sec:margin}

Equation~\eqref{eq:mkl-bound} bounds excess risk, which does not by itself imply
recovery of $\PA(Y)$. A set can have small excess risk and still be
non-invariant, and a bound on estimation error constrains which sets are accepted
only if the non-invariant sets are separated from invariance by more than that
error. We therefore state the separation as a condition.

\begin{assumption}[Invariance margin]
\label{ass:margin}
There exists $\gamma > 0$ such that every non-invariant $S$ in the candidate pool
satisfies $D(S) \ge \gamma$, while $D(\PA(Y)) = 0$.
\end{assumption}

Assumption~\ref{ass:margin} plays the role that a beta-min condition plays in
sparse regression~\cite{meinshausen2006high,wainwright2009sharp,buhlmann2011statistics}:
without it, arbitrarily weak departures from invariance are indistinguishable
from noise at any sample size.

\begin{theorem}[Recovery of the minimal invariant set]
\label{thm:recovery}
Let Assumptions~\ref{ass:modularity}--\ref{ass:cme} hold, write
$E := |\Ecal|$, and suppose the screening stage retains $\PA(Y)$ within the pool
$\Scal_{\mathrm{pool}}$ with probability at least $1-\delta_1$. Accept $S$ when
$\widehat{D}(S) \le 4E\varepsilon_n^2$, with $\varepsilon_n$ as in
Assumption~\ref{ass:cme}. If
\begin{equation}
\label{eq:margin-condition}
\gamma > 4E\big(1+\sqrt{E}\big)^2 \varepsilon_n^2 ,
\end{equation}
then with probability at least $1-\delta_1-\delta_2$ the minimal invariant set
computed from the estimated operators equals $\PA(Y)$.
\end{theorem}

Substituting the rate of Assumption~\ref{ass:cme} into
\eqref{eq:margin-condition} and combining with the screening requirement
$n \gtrsim s\log d$ gives
\begin{equation}
\label{eq:composite}
n \;\gtrsim\; \max\Big\{\, s\log d,\;
\Big(\tfrac{C\,E(1+\sqrt{E})^2\log(KE/\delta_2)}{\gamma}\Big)^{1/(2b)}
\,\Big\} .
\end{equation}
The two terms have different origins. The first comes from screening and is
logarithmic in the ambient dimension $d$. The second comes from embedding
estimation; it grows only logarithmically in the pool size $K$, but carries the
exponent $b$, and is the binding constraint whenever $b < 1/2$. The exponent
$1/(2b)$ rather than $1/b$ arises because $D(S)$ is quadratic in the operator
error. The ambient dimension therefore enters this analysis only through
screening, a point that Section~\ref{sec:exp-recovery} examines empirically and
does not fully confirm.

\begin{remark}[Multiplicity]
\label{rem:multiplicity}
Theorem~\ref{thm:recovery} is a joint statement: it requires a correct decision
on every set in the candidate pool simultaneously. With a pool of $K$ sets and
per-set error probability $\eta$, the joint statement fails with probability up
to $K\eta$, and $K$ grows combinatorially in the pool size. Finite-sample
recovery can therefore fail while every individual claim of
Section~\ref{sec:identifiability} holds. Section~\ref{sec:exp-verify} separates
these two effects by reporting per-claim rates and the joint rate separately.
\end{remark}

Two boundaries of the analysis are worth marking. Excess-risk bounds do not by
themselves imply support recovery, which is why Assumption~\ref{ass:margin} is
needed; and the sample complexity is not $O(s\log d)$, since the $n$-dependence
is inherited from embedding estimation, as \eqref{eq:composite} records.

\section{Algorithm}
\label{sec:algorithm}

The procedure has two stages, corresponding to the two terms in
\eqref{eq:composite}.

The first stage reduces the number of covariates under consideration. Each
covariate $X_j$ contributes a block of $m$ random Fourier
features~\cite{rahimi2007rff}, and we fit a group lasso over the $d$ blocks,
which for additive kernels is $\ell_1$-constrained multiple kernel
learning~\cite{bach2008consistency}. We retain the $K$ covariates with the
largest block norms, with $\lambda \asymp \sqrt{\log d / n}$ as
\eqref{eq:mkl-bound} suggests.

The second stage searches the screened pool. We enumerate subsets in order of
increasing size, test each for invariance, and stop at the first size at which some set is accepted.
Enumeration by size is used rather than a greedy search because invariance is not
monotone in $S$~\cite{peters2016icp}: a greedy pass may traverse non-invariant
intermediate sets and terminate away from the minimal one. We report two selection rules computed from the same enumeration. The
\emph{minimal} rule returns an accepted set of smallest cardinality, with ties
broken by cross-validated risk, and is the empirical counterpart of
Theorem~\ref{thm:minimal}. The \emph{intersection} rule returns the
intersection of every accepted set across all sizes, and is the empirical
counterpart of Corollary~\ref{cor:icp}. The distinction matters in finite
samples: the minimal rule stops at the smallest accepted size, so a single false
acceptance at a small size determines the output, whereas an intersection over
all sizes is affected by such an acceptance only through the variables it shares
with the other accepted sets. Restricting the intersection to a single
cardinality does not estimate the quantity in Corollary~\ref{cor:icp} and
inherits the fragility of the minimal rule.

The cost is dominated by the second stage and bounds the regime in which the
method is usable. Screening fits a group lasso over $d$ blocks of $m$ random
features at $O(dmn)$ per iteration. The enumeration then performs at most
\[
\sum_{r=1}^{s}\binom{K}{r} \;=\; O\!\big(K^{s}\big)
\]
invariance tests on a pool of size $K$, stopping earlier when a set is accepted.
One test costs $O(nm^2 + nmB)$: a cross-validated ridge fit in an $m$-dimensional
basis, and $B$ permutations of a statistic linear in $n$, the residual kernel
being applied through an explicit feature map. The procedure is thus polynomial
in $n$ and $d$ but exponential in $s$, and suits the regime it was designed for:
$d$ large, $s$ small, $K$ a small multiple of $s$. Since $K$ drives the cost
while only $\log K$ enters \eqref{eq:composite}, enlarging the pool is cheap
statistically and expensive computationally, which is the trade-off screening
manages.

\subsection{Designing the invariance test}
\label{sec:test-design}

Stage 2 requires a test of the hypothesis that $S$ is invariant, and the design
of that test is not routine in this setting. The difficulty is that interventions
shift $P_e(X_S)$, so different environments occupy different regions of covariate
space. Any statistic that does not condition exactly on $X_S$ can read that
separation as a difference between conditional laws. We therefore compare
candidate designs under identical conditions, measuring Type I error on
$\PA(Y)$, which is invariant by Assumption~\ref{ass:modularity} so that every
rejection is a false positive. Table~\ref{tab:test-design} reports level and
power for each design; Figure~\ref{fig:calibration} in
Appendix~\ref{app:testdesign} plots the levels.

\input{tables/test_design}

Appendix~\ref{app:testdesign} gives the per-design analysis. Two designs that
suggest themselves are badly miscalibrated: stratifying on the fitted value of a
pooled regression conditions on $X_S$ only through a one-dimensional summary and
reaches a Type I error of $0.493$, while the generalised covariance
measure~\cite{shah2020hardness} needs a nuisance estimate of
$\mathbb{E}[\mathbb{1}\{E=e\}\mid X_S]$ that strong interventions make close to
deterministic, and rejects almost always. The design we adopt combines a mean
model rich enough that its bias does not vary appreciably with environment --- an
explicit linear term plus random Fourier blocks at several bandwidths, with a
cross-validated ridge --- with stratification on $X_S$ itself by $k$-means, and
permutes environment labels within strata so that the null holds the covariate
design fixed. It attains Type I errors of $0.067$, $0.060$ and $0.087$ across the
three mechanisms. These values are substantially below those of the rejected
designs but remain above the nominal level in two mechanisms; we therefore treat
the test as approximately calibrated and do not claim exact finite-sample Type I
error control. Its cost is power against supersets of $\PA(Y)$, which does not
affect the reported minimal set because the enumeration accepts $\PA(Y)$ first.

\section{Experiments}
\label{sec:experiments}

\subsection{Setup}

Data come from random structural causal models over $d$ covariates with $s$
direct causes of $Y$ and three children of $Y$, observed in five environments.
Parents are drawn from the variables preceding $Y$ in the topological order and
children from those following it, so that the children are predictive of $Y$ but
cease to be invariant once their mechanisms are perturbed. Environment $0$ is
observational; the others apply sparse mechanism shifts to a random fraction of
the mechanisms, always including the children of $Y$, so that
Assumption~\ref{ass:descendant} holds, and, depending on the regime, the parents
of $Y$, which controls Assumption~\ref{ass:parent}. The mechanism of $Y$ is never
intervened on, so Assumption~\ref{ass:modularity} holds by construction and
$\PA(Y)$ is invariant as a property of the generator rather than of estimation.

Two mechanisms are used, and the distinction between them is dictated by
Assumption~\ref{ass:parent}. The \emph{monotone} mechanism is nonlinear but
unbounded, so a shift in any parent reaches $Y$ and the non-cancellation clause
holds. The \emph{saturating} mechanism is bounded, so once a parent shift moves
its argument into the flat region, shifts in the remaining parents no longer
reach $Y$: the clause fails and $\PA(Y)$ is not identifiable. We use the monotone
mechanism as the main setting and the saturating one as a controlled violation of
Assumption~\ref{ass:parent}. Appendix~\ref{app:setup} gives the mechanism
formulas, the intervention magnitudes and noise scales, and the hyperparameters
of the procedure; Appendix~\ref{app:generator} records two properties of the
generator that require care when $d$ is large.

\subsection{Population-level verification}
\label{sec:exp-verify}

\input{tables/verify_table}

The analysis of Section~\ref{sec:identifiability} makes numerical predictions
about which sets are invariant. We ask whether those predictions hold when
estimation error is small, so that a disagreement would indicate an error in the
analysis rather than in the estimator.

The study uses $d=8$ and $|\PA(Y)|=3$, so that all subsets up to size $3$ can be
enumerated exhaustively, and increase $n$ from $500$ to $2000$. For each of eight
random models we record five quantities: the rate at which $\PA(Y)$ is accepted
(C1), the fraction of strict subsets of $\PA(Y)$ rejected (C2), the fraction of
sets $\PA(Y)\cup\{Z\}$ with $Z$ a non-descendant that are accepted (C3), the
fraction of descendants of $Y$ rejected when tested alone (C4), and the joint
event that no subset smaller than $\PA(Y)$ is accepted while $\PA(Y)$ is (C5).

Each claim of the analysis makes a definite prediction about these quantities.
Proposition~\ref{prop:pa-invariant} predicts C1 near the nominal acceptance rate,
and Assumptions~\ref{ass:parent} and~\ref{ass:descendant} predict C2 and C4 equal
to $1$. The two accounts of what invariance identifies differ on C3:
Proposition~\ref{prop:superset} predicts $1$, uniqueness predicts $0$.
Theorem~\ref{thm:recovery} predicts C5 approaches $1$ as $\varepsilon_n$ falls
below the margin, and Remark~\ref{rem:multiplicity} predicts C5 lags C1--C4 at
small $n$.

Table~\ref{tab:verify} reports the measurements. C1 equals $1.00$ at every
sample size, and C2 and C4 reach $1.00$ by $n=1000$. C3 lies between $0.92$ and
$1.00$ throughout, so sets strictly containing $\PA(Y)$ are accepted as
invariant at close to every opportunity. C5 rises from $0.38$ at $n=500$ to
$0.75$ at $n=1000$ and $0.88$ at $n=2000$. Rates are computed over eight
replicates, so their resolution is limited; the variation of C3 between $n=1000$
and $n=2000$ is within that resolution.

The C3 measurements are consistent with Proposition~\ref{prop:superset} and
inconsistent with uniqueness of the invariant set. Because these measurements are
made where estimation error is small, the failure of uniqueness is a property of
the criterion rather than of the test. The behaviour of C5 relative to C1--C4 is
consistent with Remark~\ref{rem:multiplicity}: the joint statement requires tens
of simultaneously correct decisions, so it fails at small $n$ even when each
individual claim holds at a high rate, and it is recovered as $n$ grows. These
results do not establish that the assumptions hold in any particular application;
they establish that when the assumptions hold by construction, the analysis
describes the outcome.

\subsection{Recovery of the parent set}
\label{sec:exp-recovery}

Equation~\eqref{eq:composite} requires $n$ to exceed a threshold set by the
invariance margin before recovery is possible, and separately requires only
$\log d$ growth. These are two distinct predictions, so we test them with two
experiments: recovery against $n$ at fixed $d$, and recovery against $d$ at fixed
$n$.

Both experiments use $s=4$ direct causes and five environments, and compare against pooled
lasso, linear ICP~\cite{peters2016icp}, IRMv1~\cite{arjovsky2019irm}, HSIC
screening~\cite{gretton2005hsic}, and an ablation reporting the screening
output without the invariance stage. Because the procedure enumerates subsets of
a screened pool, a screening failure cannot be recovered from, so we report
screening recall separately; it upper-bounds what the invariance stage can
achieve.

\input{tables/n_table}

\begin{figure}[t]
\centering
\includegraphics[width=\columnwidth]{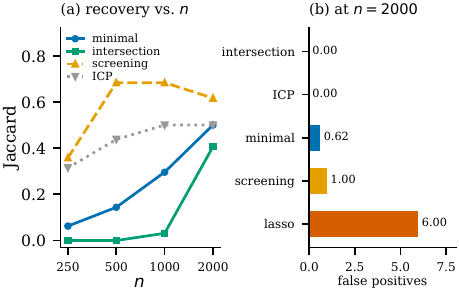}
\caption{Recovery at $d=1000$. (a) Both selection rules improve with $n$, the
threshold Theorem~\ref{thm:recovery} predicts; screening does not improve, since
it does not test invariance, and scores higher only by always reporting four
covariates. (b) False positives at the largest $n$: the intersection rule and
linear ICP admit none, matching Corollary~\ref{cor:icp}. Panel (b) is why (a)
understates the procedure, since Jaccard rewards overlap rather than the
avoidance of non-causes.}
\Description{Two panels showing recovery increasing with sample size and the
number of false positives for the compared methods at the largest sample size.}
\label{fig:recovery}
\end{figure}

Table~\ref{tab:nsweep} and Figure~\ref{fig:recovery} report recovery against
$n$ at $d=1000$. The behaviour
predicted by Theorem~\ref{thm:recovery} is a threshold: while the estimation
error exceeds the margin, non-invariant sets are accepted and neither selection
rule has reason to return $\PA(Y)$. We observe this. Recovery under the minimal
rule rises from $0.06$ at $n=250$ to $0.50$ at $n=2000$, and exact recovery from
$0.00$ to $0.38$. The mechanism is visible in the accepted sets. At $n=500$ the
test accepts on average $25$ of the $210$ four-element subsets of the screened
pool and $0.8$ of the $10$ singletons, so the minimal rule frequently stops at a
falsely accepted singleton and the intersection over all accepted sets is empty.
At $n=2000$ the same counts fall to $4.0$ and $0.0$: no singleton is accepted, so
the minimal rule cannot terminate before reaching sets of the size of $\PA(Y)$.
The transition is in the test's power, not in the identifiability analysis, which
is identical at both sample sizes.
Recovery at small $n$ is therefore not a matter of tuning; the sample size must be
large enough for the test to separate invariant from non-invariant sets.

Comparing against the screening-only ablation separates two different objectives.
At $n=2000$ that ablation attains a higher Jaccard index than either selection
rule ($0.62$ against $0.50$ and $0.41$), because it always reports four
covariates and so cannot under-select; it also returns one false positive per
selection on average, with no guarantee against doing so. The intersection rule
returns none, matching Corollary~\ref{cor:icp}, as does linear ICP. The
invariance stage therefore purchases a constraint on false positives rather than
an improvement in overlap, and a comparison on overlap alone does not measure
what it provides. Pooled lasso makes the contrast plain: perfect recall by
selecting broadly, at six false positives per selection.

The guarantee has a limit worth recording. The intersection rule returns $0.38$
false positives per selection at $n=1000$,
which the population statement of Corollary~\ref{cor:icp} would not allow. The
corollary holds because $\PA(Y)$ is one of the intersected sets; when the test
rejects $\PA(Y)$ in a given replicate, the accepted family need not contain it
and the intersection need not lie inside it. The guarantee is thus conditional on
$\PA(Y)$ being accepted, which is a property of the test rather than of the
identifiability analysis.

Turning to the second prediction, screening recall behaves as
\eqref{eq:composite} requires, falling only from
$0.88$ at $d=200$ to $0.75$ at $d=2000$. Recovery itself is less stable and is
not monotone in $d$. Appendix~\ref{app:pool} gives the table, together with a
follow-up experiment locating the reason in the composition of the screened pool
rather than in its size; we draw no conclusion from this sweep about the
$d$-dependence of the second stage.

\subsection{Ablation on the parent-activation assumption}
\label{sec:exp-ablation}

A final experiment varies the conditions rather than the method, moving the
parent-activation regime between Assumption~\ref{ass:parent} holding, holding
weakly, and failing. Where it fails, $\PA(Y)$ is not identifiable from
invariance, so the appropriate outcome is degradation rather than recovery.
That is what we observe, and the two stages separate cleanly: recovery under the
intersection rule falls from $0.41$ to $0.15$ to $0.03$, returning close to the
empty set rather than an incorrect one, while the screening-only ablation is
comparatively unaffected, moving only from $0.62$ to $0.57$, as expected of a
criterion that measures association and does not depend on
Assumption~\ref{ass:parent}. Appendix~\ref{app:diversity} gives the full table.

\section{Discussion}
\label{sec:discussion}

The operator formulation restates invariant causal prediction rather than
extending its identification power. Its value is representational: it requires
no parametric conditional model and expresses departure from invariance as a
single scalar. By Proposition~\ref{prop:ci}, it identifies exactly what the
conditional independence $Y \indep E \mid X_S$ identifies: a minimal set rather
than a unique one, and only under faithfulness-type environment conditions.

The binding practical constraint is estimation rather than dimension. Sparsity
provides the favourable dependence on $d$ in \eqref{eq:composite}, whereas
embedding estimation controls the generally slower dependence on $n$. Below the
sample size needed to resolve the margin, the minimal rule can be determined by
one false acceptance, while the intersection rule shrinks towards the empty set
as false acceptances accumulate. Minimality is therefore the sharper population
statement, but intersection is the more defensible estimator when the margin is
unknown because it fails conservatively.

The calibration study exposes a related tension. Large environment differences
make invariance informative but can make the environment indicator nearly
deterministic given the covariates. Tests that estimate that indicator, including
the generalised covariance measure, can therefore be least reliable in this
regime.

Several limitations bound these results. The non-cancellation clauses in
Assumptions~\ref{ass:descendant} and~\ref{ass:parent} are not empirically
checkable, and recovery is conditional on screening success. Under strong
covariate shift the permutation test is only approximately calibrated, as
Section~\ref{sec:test-design} quantifies. Finally, the benchmarks are synthetic:
known causal structure enables direct verification, but establishes neither
performance on real covariates nor utility in a particular domain. A real-data
benchmark would need interventions with sufficient common support
(Remark~\ref{rem:support}); our attempts instead lost calibration because of the
benchmark design. The experiments therefore validate the analysis internally,
not its external applicability.

\section{Conclusion}

We asked what environment invariance identifies and what identification costs.
Invariance identifies the direct causes of $Y$ only up to minimality: every
superset of $\PA(Y)$ that omits descendants of $Y$ is invariant as well, so the
parent set is the smallest invariant set rather than the only one, and even that
requires two separate conditions on the environments. The cost of establishing
this from data is governed by conditional mean embedding estimation, not by the
ambient dimension, which enters only through a logarithmic screening term.
Experiments run where estimation error is small reproduce each of these
predictions, including the failure of uniqueness. For practice, the useful
consequence is a division of labour within the framework: screening supplies the
tolerance to high dimension, the invariance criterion supplies identification up
to minimality, and the conditions on the environments determine whether that
minimal set is the causal one.

\bibliographystyle{ACM-Reference-Format}
\bibliography{refs}

\clearpage
\appendix
\input{appendix_camera_ready}

\end{document}

%% file: figures/lattice.tex
\begin{tikzpicture}[
  x=0.9cm, y=0.95cm,
  every node/.style={font=\scriptsize, align=center},
  inv/.style={draw=green!50!black, fill=green!10, rounded corners=1.6pt,
              inner sep=2.2pt, line width=0.5pt},
  pa/.style={draw=green!45!black, fill=green!30, rounded corners=1.6pt,
             inner sep=2.6pt, line width=1.1pt},
  non/.style={draw=red!70!black, fill=red!7, rounded corners=1.6pt,
              inner sep=2.2pt, line width=0.5pt},
  link/.style={draw=gray!55, line width=0.5pt, shorten >=1.5pt, shorten <=1.5pt},
  key/.style={font=\scriptsize, text=black!75, anchor=west}
]
\node[inv] (z2) at ( 0.15, 3.15) {$\PA(Y)\cup\{Z,Z'\}$};
\node[inv] (z1) at (-1.15, 1.95) {$\PA(Y)\cup\{Z\}$};
\node[non] (d1) at ( 1.65, 1.95) {$\PA(Y)\cup\{D\}$};
\node[pa]  (pa) at ( 0.15, 0.75) {$\PA(Y)$};
\node[non] (s1) at (-1.20,-0.45) {$\{X_1\}$};
\node[non] (s2) at ( 1.35,-0.45) {$\{X_1,X_2\}$};

\draw[link] (s1) -- (pa);   \draw[link] (s2) -- (pa);
\draw[link] (pa) -- (z1);   \draw[link] (pa) -- (d1);
\draw[link] (z1) -- (z2);

\draw[gray!65, line width=0.4pt] (pa.east) -- ++(0.55,0)
      node[key, font=\scriptsize\itshape, inner sep=1pt] {minimal};

\node[inv, minimum width=5pt, minimum height=5pt, inner sep=1.6pt]
      (li) at (-1.55,-1.55) {};
\node[key] at (-1.35,-1.55) {invariant};
\node[non, minimum width=5pt, minimum height=5pt, inner sep=1.6pt]
      (ln) at ( 0.75,-1.55) {};
\node[key] at ( 0.95,-1.55) {not invariant};
\end{tikzpicture}

%% file: tables/test_design.tex
\begin{table*}[t]
\centering
\small
\caption{Invariance-test designs under identical conditions ($d=10$, $n=2000$, five environments, shift $2.0$; 150 replicates for level, 100 for power). Level is measured on $\PA(Y)$, invariant by Assumption~\ref{ass:modularity}, so every rejection is a false positive; nominal level $0.05$.}
\label{tab:test-design}
\begin{tabular}{lccccc}
\toprule
& \multicolumn{3}{c}{Type I error} & \multicolumn{2}{c}{Power} \\
\cmidrule(lr){2-4}\cmidrule(lr){5-6}
Design & linear & monot. & satur. & $S\subsetneq\PA(Y)$ & $\PA(Y)\cup\{D\}$ \\
\midrule
Fitted-value strata, single-scale model & 0.493 & 0.500 & 0.240 & 0.980 & 0.810 \\
Generalised covariance measure & 1.000 & 1.000 & 0.973 & 1.000 & 1.000 \\
Rich model, no stratification & 0.060 & 0.020 & 0.160 & 1.000 & 0.860 \\
Rich model, covariate strata, local adjustment & 0.040 & 0.060 & 0.080 & 1.000 & 0.160 \\
Rich model, covariate strata \textbf{(ours)} & 0.067 & 0.060 & 0.087 & 1.000 & 0.280 \\
\bottomrule
\end{tabular}
\end{table*}

%% file: tables/verify_table.tex
\begin{table}[t]
\centering\small
\caption{Population-level verification at $d=8$, eight replicates. C1--C4 are per-claim rates; C5 is the joint statement the selection procedure relies on. A C3 value near $1$ is inconsistent with uniqueness of the invariant set.}
\label{tab:verify}
\begin{tabular}{rccccc}
\toprule
$n$ & C1 & C2 & C3 & C4 & C5 \\
& $\PA(Y)$ inv. & subsets rej. & supersets inv. & desc. rej. & joint \\
\midrule
500 & 1.00 & 0.92 & 0.96 & 0.94 & 0.38 \\
1000 & 1.00 & 1.00 & 1.00 & 1.00 & 0.75 \\
2000 & 1.00 & 1.00 & 0.92 & 1.00 & 0.88 \\
\midrule
\textit{theory} & $\ge 0.95$ & $1.00$ & $1.00$ & $1.00$ & $\to 1$ \\
\bottomrule
\end{tabular}
\end{table}

%% file: tables/n_table.tex
\begin{table*}[t]
\centering\small
\caption{Recovery against sample size at $d=1000$, $s=4$, five environments, eight replicates. Theorem~\ref{thm:recovery} predicts poor recovery until the estimation error falls below the margin. The right-hand block reports precision, recall and false positives at the largest $n$: the invariance stage controls false positives (Corollary~\ref{cor:icp}), not overlap. Screening recall bounds what any later stage can recover.}
\label{tab:nsweep}
\begin{tabular}{lcccc@{\hskip 8pt}ccc}
\toprule
& \multicolumn{4}{c}{Jaccard index} & \multicolumn{3}{c}{at $n=2000$} \\
\cmidrule(lr){2-5}\cmidrule(lr){6-8}
Method & $n=250$ & $n=500$ & $n=1000$ & $n=2000$ & prec. & rec. & f.p. \\
\midrule
Kernel ICP (minimal) & 0.06 & 0.14 & 0.30 & 0.50 & 0.59 & 0.53 & 0.62 \\
Kernel ICP (intersection) & 0.00 & 0.00 & 0.03 & 0.41 & 0.50 & 0.41 & 0.00 \\
MKL only (no invariance) & 0.36 & 0.68 & 0.68 & 0.62 & 0.75 & 0.75 & 1.00 \\
ICP (linear) & 0.31 & 0.44 & 0.50 & 0.50 & 0.50 & 0.50 & 0.00 \\
Lasso (pooled) & 0.40 & 0.40 & 0.40 & 0.40 & 0.40 & 1.00 & 6.00 \\
HSIC screening & 0.00 & 0.00 & 0.00 & 0.00 & 0.00 & 0.00 & 4.00 \\
IRMv1 (linear) & 0.00 & 0.00 & 0.00 & 0.00 & 0.00 & 0.00 & 4.00 \\
\midrule
\emph{screening recall} & 0.56 & 0.84 & 0.81 & 0.84 & & & \\
\bottomrule
\end{tabular}
\end{table*}

%% file: appendix_camera_ready.tex
\section{Proofs for Section~\ref{sec:criterion}}
\label{app:criterion}

\begin{proof}[Proof of Proposition~\ref{prop:ci}]
Write $\mu^{(e)}_{Y\mid X_S=x} := \CME{Y}{X_S}^{(e)} k_{\Xcal_S}(x,\cdot) \in
\Hcal_\Ycal$ for the conditional mean embedding of $P_e(Y \mid X_S = x)$.

($\Rightarrow$) Suppose $\CME{Y}{X_S}^{(e)} = \CME{Y}{X_S}^{(e')}$ as operators.
Applying both sides to $k_{\Xcal_S}(x,\cdot)$ gives
$\mu^{(e)}_{Y\mid X_S=x} = \mu^{(e')}_{Y\mid X_S=x}$ for every $x$, hence for
$P_e$-almost every $x$; by mutual absolute continuity the exceptional set is null
under every environment. Because $k_\Ycal$ is characteristic, the mean embedding
is injective on Borel probability measures~\cite{sriperumbudur2010hilbert}, so
$P_e(Y \mid X_S = x) = P_{e'}(Y \mid X_S = x)$ for almost every $x$. As this
holds for every pair $e, e'$, the conditional law of $Y$ given $X_S$ does not
depend on the environment, which is precisely $Y \indep E \mid X_S$.

($\Leftarrow$) If $Y \indep E \mid X_S$ then $P_e(Y\mid X_S = x)$ is the same
measure for all $e$ at almost every $x$, so the embeddings agree and therefore
$\CME{Y}{X_S}^{(e)}$ and $\CME{Y}{X_S}^{(e')}$ agree on
$\{k_{\Xcal_S}(x,\cdot) : x \in \operatorname{supp} P(X_S)\}$. The span of this
set is dense in $\Hcal_{\Xcal_S}$ by the reproducing property, and both operators
are bounded, so they agree on $\Hcal_{\Xcal_S}$.
\end{proof}

\begin{proof}[Proof of Proposition~\ref{prop:collapse}]
(i) The statement $Y \indep E \mid \Phi(X)$ is a statement about the
$\sigma$-algebra $\sigma(\Phi(X))$ only: conditional independence given a random
element depends on that element solely through the $\sigma$-algebra it
generates. If $w$ is a bimeasurable bijection then
$\sigma(w\circ\Phi(X)) = \sigma(\Phi(X))$, so the criterion is unchanged.

(ii) If $w$ is not injective on the range of $\Phi$ then
$\sigma(w\circ\Phi(X)) \subsetneq \sigma(\Phi(X))$. Conditional independence is
not monotone in the conditioning $\sigma$-algebra in either direction, so
coarsening can create or destroy invariance; part (iii) exhibits the extreme
case.

(iii) If $\Phi \equiv c$ is constant then $\sigma(\Phi(X))$ is the trivial
$\sigma$-algebra and $Y \indep E \mid \Phi(X)$ reduces to $Y \indep E$, i.e.\ to
$P_e(Y) = P_{e'}(Y)$ for all $e,e'$. Hence if some environment changes the
marginal law of $Y$, no constant representation is invariant.
\end{proof}

\begin{remark}
Part (iii) is what defeats the ``absorb $w$ into $\Phi$'' objection to bilevel
invariance objectives. Setting $\Phi' = w\circ\Phi$ does not make the constraint
vacuous, because the constraint is a property of the $\sigma$-algebra generated
by the representation, not of the pair $(\Phi,w)$. In
Section~\ref{sec:identifiability} the objection cannot even be raised, since
identifiability is stated over variable subsets, for which no reparameterisation
exists.
\end{remark}

\subsection*{Reparameterisation of a representation}

\begin{proposition}[Reparameterisation]
\label{prop:collapse}
Define invariance of a measurable $\Phi$ as $Y \indep E \mid \Phi(X)$.
\begin{enumerate}
\item[(i)] The criterion depends on $\Phi$ only through the
  $\sigma$-algebra it generates. If $w$ is a bimeasurable bijection, then
  $\Phi$ is invariant if and only if $w \circ \Phi$ is.
\item[(ii)] If $w$ is not injective on the range of $\Phi$, then
  $\sigma(w \circ \Phi(X)) \subsetneq \sigma(\Phi(X))$, and invariance of one does
  not imply invariance of the other in either direction.
\item[(iii)] A constant $\Phi$ is invariant if and only if $Y \indep E$, that is,
  if and only if the marginal law of $Y$ is the same in every environment.
\end{enumerate}
\end{proposition}

Part (iii) identifies when the degenerate representation satisfies the criterion:
exactly when the environments leave the marginal law of $Y$ unchanged. In the
regime where invariance is informative, some environment shifts that law, and the
constant representation is then not invariant. We note that the question does not
arise for the results in Section~\ref{sec:identifiability}, which are stated over
covariate subsets, for which no reparameterisation is available.

\section{Proofs for Section~\ref{sec:identifiability}}
\label{app:identifiability}

\begin{proof}[Proof of Proposition~\ref{prop:pa-invariant}]
By Assumption~\ref{ass:modularity} the structural assignment
$Y := f_Y(\PA(Y), U_Y)$ and the law of $U_Y$ are identical in every environment,
and $U_Y$ is independent of $\PA(Y)$ in every environment. Hence for every $e$
and every measurable $A$,
\[
P_e\big(Y \in A \mid X_{\PA(Y)} = x\big)
= P\big(f_Y(x, U_Y) \in A\big),
\]
which does not depend on $e$. Applying Proposition~\ref{prop:ci} in the
direction ($\Leftarrow$) gives $\PA(Y) \in \Scal_{\mathrm{inv}}$.
\end{proof}

\begin{proof}[Proof of Proposition~\ref{prop:superset}]
Let $S \supseteq \PA(Y)$ with $S \cap \DE(Y) = \emptyset$ and put
$R := S \setminus \PA(Y)$, so $R$ consists of non-descendants of $Y$. The local
Markov property for the SCM in environment
$e$~\cite{pearl2009causality,peters2017elements} gives that $Y$ is conditionally
independent of its non-descendants given its parents, hence
$Y \indep X_R \mid X_{\PA(Y)}$ under $P_e$ and therefore
\[
P_e(Y \mid X_S) = P_e(Y \mid X_{\PA(Y)}, X_R) = P_e(Y \mid X_{\PA(Y)}).
\]
Interventions may alter the graph over $X$, but they do not add edges into $Y$
under Assumption~\ref{ass:modularity}, so $\PA(Y)$ remains the parent set of $Y$
in every environment and the display holds for each $e$. By
Proposition~\ref{prop:pa-invariant} the right-hand side does not depend on $e$,
so neither does the left-hand side, and $S \in \Scal_{\mathrm{inv}}$ by
Proposition~\ref{prop:ci}.
\end{proof}

\begin{proof}[Proof of Corollary~\ref{cor:icp}]
By Proposition~\ref{prop:pa-invariant}, $\PA(Y) \in \Scal_{\mathrm{inv}}$.
An intersection over a family is contained in each of its members, so
$\bigcap_{S\in\Scal_{\mathrm{inv}}} S \subseteq \PA(Y)$.
\end{proof}

\begin{proof}[Proof of Theorem~\ref{thm:minimal}]
We show
\begin{equation}
\label{eq:inv-family-proof}
\Scal_{\mathrm{inv}} \;=\; \big\{\, S \;:\; \PA(Y) \subseteq S,\;
S \cap \DE(Y) = \emptyset \,\big\}.
\end{equation}

($\supseteq$) is Proposition~\ref{prop:superset}.

($\subseteq$) Let $S \in \Scal_{\mathrm{inv}}$. If $S \not\supseteq \PA(Y)$ then
Assumption~\ref{ass:parent} supplies $e,e'$ with
$P_e(Y \mid X_S) \neq P_{e'}(Y \mid X_S)$, so by Proposition~\ref{prop:ci} the
operators differ and $S \notin \Scal_{\mathrm{inv}}$, a contradiction; hence
$S \supseteq \PA(Y)$. If $S \cap \DE(Y) \neq \emptyset$ then
Assumption~\ref{ass:descendant} gives the same contradiction; hence
$S \cap \DE(Y) = \emptyset$. This proves~\eqref{eq:inv-family-proof}.

The right-hand side of~\eqref{eq:inv-family-proof} is the collection of all sets
sandwiched between $\PA(Y)$ and the complement of $\DE(Y)$. It is nonempty
(it contains $\PA(Y)$, since $\PA(Y) \cap \DE(Y) = \emptyset$ by acyclicity)
and closed upward within that complement. Its unique minimal element under
inclusion is $\PA(Y)$, and since every member contains $\PA(Y)$ while $\PA(Y)$
is itself a member, $\bigcap_{S \in \Scal_{\mathrm{inv}}} S = \PA(Y)$.
\end{proof}

\begin{proof}[Proof of Corollary~\ref{cor:fixedpoint}]
By~\eqref{eq:inv-family-proof}, let $S \in \Scal_{\mathrm{inv}}$.

If $S \supsetneq \PA(Y)$, pick $j \in S \setminus \PA(Y)$. Then
$S\setminus\{j\} \supseteq \PA(Y)$ and $(S\setminus\{j\}) \cap \DE(Y) =
\emptyset$, so $S \setminus \{j\} \in \Scal_{\mathrm{inv}}$
by~\eqref{eq:inv-family-proof}, whence $j \notin \Psi(S)$ and $\Psi(S) \neq S$. So no
strict superset of $\PA(Y)$ is a fixed point.

If $S = \PA(Y)$, then for every $j \in S$ the set $S \setminus \{j\}$ fails to
contain $\PA(Y)$, so $S\setminus\{j\} \notin \Scal_{\mathrm{inv}}$
by~\eqref{eq:inv-family-proof}; thus every $j \in S$ lies in $\Psi(S)$ and
$\Psi(\PA(Y)) = \PA(Y)$. Hence $\PA(Y)$ is the unique fixed point.
\end{proof}

\begin{remark}[Where each assumption is used]
Assumption~\ref{ass:modularity} alone yields Propositions~\ref{prop:pa-invariant}
and~\ref{prop:superset} and Corollary~\ref{cor:icp}. Assumption~\ref{ass:parent}
is used only for the inclusion $S \supseteq \PA(Y)$, and
Assumption~\ref{ass:descendant} only for $S \cap \DE(Y) = \emptyset$. The two
are therefore separately necessary for~\eqref{eq:inv-family-proof}: dropping
Assumption~\ref{ass:parent} admits invariant strict subsets of $\PA(Y)$, and
dropping Assumption~\ref{ass:descendant} admits invariant sets containing
descendants. Section~\ref{sec:experiments} exhibits both failures.
\end{remark}

\subsection*{The fixed-point formulation}

Minimality can be expressed as a fixed-point property, which is the form used by
the algorithm of Section~\ref{sec:algorithm}. The operator must be defined on the
invariant family, since it asks which elements of a set are indispensable for
invariance.

\begin{definition}[Indispensability operator]
\label{def:psi}
For $S \in \Scal_{\mathrm{inv}}$, let
$\Psi(S) := \{\, j \in S : S \setminus \{j\} \notin \Scal_{\mathrm{inv}} \,\}$.
\end{definition}

A set is a fixed point of $\Psi$ when no single deletion preserves invariance,
that is, when it is locally minimal in $\Scal_{\mathrm{inv}}$. Under the
conditions of Theorem~\ref{thm:minimal}, local and global minimality coincide.

\begin{corollary}[Fixed points]
\label{cor:fixedpoint}
Under the assumptions of Theorem~\ref{thm:minimal}, $\PA(Y)$ is the unique fixed
point of $\Psi$ in $\Scal_{\mathrm{inv}}$.
\end{corollary}

\begin{remark}[Invariance is a property of a set]
\label{rem:marginal}
A per-variable criterion, retaining $X_j$ when the conditional operator given
$X_j$ alone is invariant, is not equivalent to Definition~\ref{def:invariance}.
For $j \in \PA(Y)$ the marginal conditional $P_e(Y \mid X_j)$ generally does vary
across environments, because the remaining parents' distributions shift.
Conversely, a variable outside $\PA(Y)$ can have an invariant marginal
conditional. An argument that establishes invariance of $\CME{Y}{X_{\PA(Y)}}$ and
then draws a conclusion about each $\CME{Y}{X_j}$ separately is therefore
incomplete.
\end{remark}

\section{Proofs for Section~\ref{sec:complexity}}
\label{app:complexity}

\begin{proof}[Proof of Lemma~\ref{lem:rademacher}]
For the upper bound, Jensen's inequality and independence of the Rademacher
variables give
\begin{align*}
\mathbb{E}_\sigma \Big\| \sum_i \sigma_i k(x_i,\cdot)\Big\|_\Hcal
&\le \Big( \mathbb{E}_\sigma \Big\| \sum_i \sigma_i k(x_i,\cdot)
   \Big\|^2_\Hcal \Big)^{1/2}\\
&= \Big(\sum_i k(x_i,x_i)\Big)^{1/2} = \sqrt{\operatorname{Tr}K},
\end{align*}
since the cross terms vanish. Dividing by $n$ gives the claim. For the lower
bound, the Khintchine--Kahane inequality in a Hilbert space with the optimal
constant~\cite{latala1994khinchin} gives
\[
\mathbb{E}\Big\|\sum_i \sigma_i v_i\Big\|
\;\ge\; 2^{-1/2}\Big(\mathbb{E}\Big\|\sum_i\sigma_i v_i\Big\|^2\Big)^{1/2},
\]
and the same computation identifies the right-hand side as
$\sqrt{\operatorname{Tr}K/2}$.
\end{proof}

\begin{remark}
Under $k(x,x)\le 1$ we have $\operatorname{Tr}K \le n$, so the bound reads
$\widehat{\mathfrak{R}}_n(B_\Hcal) \le n^{-1/2}$. The variant
$\sqrt{\operatorname{Tr}K/n}$ that appears in some treatments is off by a factor
$\sqrt{n}$ and is vacuous whenever $\operatorname{Tr}K \asymp n$, since it does
not decay.
\end{remark}

\begin{proof}[Proof of Theorem~\ref{thm:recovery}]
Write $E := |\Ecal|$, abbreviate $C_e := \CME{Y}{X_S}^{(e)}$,
$\bar C := E^{-1}\sum_e C_e$, and similarly for the estimates. Assume the event
$\Omega$ on which the screening step retains $\PA(Y)$ and
$\|\widehat{C}_e - C_e\|_{\HS} \le \varepsilon$ for every $e$ and every $S$ in
the candidate pool; $P(\Omega) \ge 1 - \delta_1 - \delta_2$. On $\Omega$,
$\|\widehat{\bar C} - \bar C\|_\HS \le \varepsilon$ by the triangle inequality,
so
\begin{equation}
\label{eq:pert}
\big| \|\widehat C_e - \widehat{\bar C}\|_\HS - \|C_e - \bar C\|_\HS \big|
\;\le\; 2\varepsilon
\qquad \text{for every } e .
\end{equation}

Consider the procedure that accepts $S$ iff
$\widehat{D}(S) := \sum_e \|\widehat C_e - \widehat{\bar C}\|_\HS^2 \le \tau$,
with threshold $\tau := 4E\varepsilon^2$.

\emph{Invariant sets are accepted.} If $D(S) = 0$ then $\|C_e - \bar C\|_\HS = 0$
for every $e$, so by~\eqref{eq:pert} $\|\widehat C_e - \widehat{\bar C}\|_\HS
\le 2\varepsilon$ and $\widehat D(S) \le 4E\varepsilon^2 = \tau$.

\emph{Non-invariant sets are rejected.} If $D(S) \ge \gamma$ then, since the sum
has $E$ terms, some $e^\star$ satisfies
$\|C_{e^\star} - \bar C\|_\HS \ge \sqrt{\gamma/E}$. By~\eqref{eq:pert},
\[
\widehat D(S) \;\ge\; \|\widehat C_{e^\star} - \widehat{\bar C}\|^2_\HS
\;\ge\; \Big(\sqrt{\gamma/E} - 2\varepsilon\Big)_+^2 .
\]
This exceeds $\tau = 4E\varepsilon^2$ as soon as
$\sqrt{\gamma/E} > 2\varepsilon + 2\varepsilon\sqrt{E}$, i.e.\ as soon as
\begin{equation}
\label{eq:margin-app}
\gamma \;>\; 4E\big(1+\sqrt{E}\big)^2 \varepsilon^2 ,
\end{equation}
which is the hypothesis of the theorem.

Hence on $\Omega$ the accepted family is exactly $\Scal_{\mathrm{inv}}$
restricted to the candidate pool. Since screening retained $\PA(Y)$,
Theorem~\ref{thm:minimal} applies within the pool and its unique minimal element
is $\PA(Y)$. Therefore the computed minimal invariant set equals $\PA(Y)$ on
$\Omega$, which has probability at least $1 - \delta_1 - \delta_2$.
\end{proof}

\begin{remark}[Reading the rate]
Substituting a CME rate $\varepsilon_n \asymp \sqrt{\log(|\text{pool}|/\delta)}\,n^{-b}$
of the kind established in~\cite{li2022optimal,li2024towards}
into~\eqref{eq:margin-app} and solving for $n$ gives the second term
of~\eqref{eq:composite}. The exponent $1/(2b)$ rather than $1/b$ arises because
the discrepancy is quadratic in the operator error. With $b = 1/2$ this is the
familiar $n \gtrsim \gamma^{-1}\log(d/\delta)$; for $b < 1/2$, which is the
typical case, it is strictly worse, and this is the sense in which CME
estimation, not dimension, is the binding constraint.
\end{remark}

\section{Composition of the screened pool}
\label{app:pool}

\input{tables/main_table}

Section~\ref{sec:exp-recovery} reports that recovery falls with $d$ faster than
reduced screening recall accounts for. Two explanations are available. Under
\emph{multiplicity}, a larger $d$ means more candidate subsets and so more
opportunities for a false acceptance, to which the minimal rule is sensitive.
Under \emph{composition}, a larger $d$ supplies covariates that survive screening
while being hard to separate from the parents by invariance. The two differ in
what they predict when the pool size is varied at fixed $d$.

Multiplicity does not account for the effect. Enlarging the pool at $d=1000$ from
$6$ to $10$ to $14$ candidates improves recovery, from $0.03$ to $0.47$ to
$0.59$, tracking screening recall ($0.46$, $0.83$, $0.96$), even though the number
of subsets tested grows from $56$ to $1470$.

Composition is implicated instead. At a fixed pool size of $10$ and equal
screening recall of $0.83$, recovery is $0.16$ at $d=200$ and $0.47$ at $d=1000$,
with $15.0$ and $6.5$ falsely accepted four-element sets respectively; restricting
to replicates in which screening retained all four parents gives $27.0$ and
$5.0$. The pool at $d=200$ is harder for the invariance test than the pool at
$d=1000$, at equal size and equal recall.

This narrows the explanation without completing it. The difficulty of the second
stage is governed by which covariates enter the pool rather than by how many, and
pool composition varies with $d$ non-monotonically in our generator.

\section{The invariance-test designs}
\label{app:testdesign}

\begin{figure}[t]
\centering
\includegraphics[width=\columnwidth]{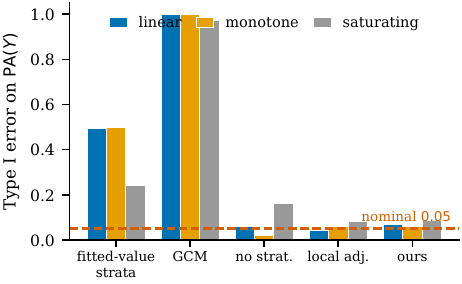}
\caption{Type I error of each test design, measured on $\PA(Y)$, so that every
rejection is a false positive. Bars show the three mechanisms; the dashed line is
the nominal level. The first two designs, the ones a reader would reach for
first, are far above it.}
\Description{Grouped bars comparing Type I error across five test designs and
three mechanisms, with a dashed horizontal line marking the nominal level.}
\label{fig:calibration}
\end{figure}

Section~\ref{sec:test-design} summarises the comparison of test designs.
This appendix gives the per-design analysis, and
Figure~\ref{fig:calibration} plots the levels.

A pooled kernel ridge regression of $Y$ on $X_S$, followed by stratification on
the fitted value and comparison of residuals across environments within strata,
conditions on $X_S$ only through a one-dimensional summary. Equal fitted values
need not correspond to nearby $x$, so environments remain imbalanced within a
stratum and the model's approximation bias varies with environment. The measured
Type I error is $0.493$ under a linear mechanism and $0.500$ under a monotone
one, against a nominal $0.05$.

The generalised covariance measure~\cite{shah2020hardness} is designed for
conditional independence testing: it regresses both $g(Y)$ and the environment
indicators on $X_S$, cross-fits, and tests whether the residual covariances
vanish. Its validity requires the product of the two regression errors to be
$o(n^{-1/2})$. Here the second regression estimates
$\mathbb{E}[\mathbb{1}\{E=e\}\mid X_S]$, and strong interventions make $E$ close
to a deterministic function of $X_S$, which is the failure of overlap described
in Remark~\ref{rem:support}. The regression is then near-degenerate and its error
does not vanish. The measured Type I error is $1.000$, and reducing the ridge
parameter from $10^{-2}$ to $10^{-6}$ lowers it only to $0.133$. This is consistent with the
interpretation that the failure is due to lack of overlap rather than to
regularisation, since the regime in which invariance is informative is exactly
the regime in which environments are well separated.

Making the mean model rich enough that its bias does not vary appreciably with
environment removes most of the error. We regress $Y$ on an explicit linear term
together with random Fourier blocks~\cite{rahimi2007rff} at several
bandwidths, with a
cross-validated ridge parameter. Under the linear and monotone mechanisms this
alone suffices: the unstratified test attains Type I errors of $0.060$ and
$0.020$, and stratification is unnecessary. Under the saturating mechanism it
does not, reaching $0.160$, because the residual spread then varies with $x$
while the environments occupy different regions of covariate space.

We therefore combine the rich mean model with stratification on $X_S$ itself by
$k$-means, which makes the model's residual bias common to all environments
within a stratum. Kernel mean embeddings of the residual are compared across
environments within strata, and environment labels are permuted within strata so
that the null distribution holds the covariate design fixed. Using an explicit
feature map for the residual kernel makes the statistic linear rather than
quadratic in $n$. Across the three mechanisms this design attains Type I errors of $0.067$,
$0.060$ and $0.087$. It is better calibrated under saturation than the
unstratified test, at a cost in power that the next paragraph quantifies.

Because strata are tight but not singletons, residual variation explained by
$X_S$ within a stratum remains, and removing it by a within-stratum linear
adjustment is a natural refinement. It lowers the Type I error further, but it
also removes signal: when environments are separated inside a stratum, a genuine
difference between conditional laws is partly collinear with the local linear
trend. Power against $\PA(Y)\cup\{D\}$ falls from $0.280$ to $0.160$, so we do
not use the adjustment. The trade-off is not specific to this implementation.
Separating a difference in conditional laws from a difference in covariate
distributions requires overlap, and no statistic can supply overlap that the
design lacks.

The last column quantifies what stratification costs. Every design rejects a
descendant $D$ when tested alone, but against $\PA(Y)\cup\{D\}$ the stratified
designs reach only $0.280$, whereas the unstratified test reaches $0.860$.
Conditioning on the full parent set leaves little residual variation for $D$ to
explain, and stratification reduces the effective sample size available to detect
what remains. The choice between the two designs is therefore a trade-off rather
than a dominance relation, and we resolve it in favour of calibration for two
reasons. Calibration under an unknown mechanism shape cannot be checked from
data, whereas the power cost is confined to supersets of $\PA(Y)$; and the
selection procedure enumerates by increasing size, so it accepts $\PA(Y)$ before
reaching any superset. The consequence is that the accepted family is not fully
resolved above $\PA(Y)$, which does not affect the reported minimal set.

\section{Parent-activation ablation}
\label{app:diversity}

\input{tables/diversity_table}

The final experiment varies the conditions rather than the method, and so tests
Assumption~\ref{ass:parent} directly. The parent-activation regime moves between
the assumption holding (every environment perturbs every parent), holding weakly
(each parent is perturbed in one environment), and failing (parents are never
perturbed). Where it fails, $\PA(Y)$ is not identifiable from invariance, so the
appropriate outcome is not recovery but degradation; a method that still reported
$\PA(Y)$ would be reporting an artefact of the procedure.

This is what we observe. Recovery under the minimal rule falls from $0.50$ when
the assumption holds, to $0.32$ under weak activation, to $0.27$ when it fails.
The intersection rule shows the pattern more sharply, falling from $0.41$ to
$0.15$ to $0.03$: as the assumption is withdrawn it returns close to the empty
set rather than an incorrect set, and its false-positive count stays at or below
$0.12$ throughout. Linear ICP~\cite{peters2016icp} behaves the same way, halving from $0.50$ to
$0.25$. The screening-only ablation is comparatively unaffected, moving from
$0.62$ to $0.57$, which is expected: screening measures association and does not
depend on Assumption~\ref{ass:parent}. The contrast between the two indicates
that the degradation is attributable to the invariance stage losing its
identifying condition rather than to a loss of signal in the data.

\section{Conditioning of the synthetic generator}
\label{app:generator}

Two properties of the generator require care when $d$ is large. First, the edge
probability is set to $2/d$ rather than held constant, so that the expected
in-degree does not grow with $d$. With a constant edge probability each variable
sums $O(d)$ parents, the propagated variance compounds down the topological
order, and at $d=2000$ we observed covariate standard deviations exceeding
$10^9$; the covariates are then numerically degenerate rather than merely
high-dimensional. Second, each column of the coefficient matrix is normalised, so
that a variable's parental contribution does not dominate its own noise, which is
required for the variance to remain bounded with depth even at constant
in-degree. Neither adjustment changes the causal structure, and both are needed
for the covariates to remain comparable across $d$.

\section{Experimental setup: details}
\label{app:setup}

This section records the quantities omitted from Section~\ref{sec:experiments}.

Writing $z$ for the linear combination of the parents, the two mechanisms are
$f_Y(z) = z + \tfrac12\sin 2z$ in the monotone case and
$f_Y(z) = \tanh z + \tfrac12\sin 2z$ in the saturating one. The first is
unbounded, so a shift in any parent reaches $Y$ at any magnitude; the second is
bounded, which is how the non-cancellation clause of
Assumption~\ref{ass:parent} is made to fail while everything else is held fixed.

An intervention on a covariate shifts its mean by $\pm 2$, with the sign drawn
independently in each environment, and inflates its noise scale in proportion to
the shift. Interventions are applied to a random $10\%$ of mechanisms in each
non-observational environment, in addition to the children and, where the regime
requires it, the parents of $Y$. The noise on $Y$ has standard deviation $0.5$
throughout and is Gaussian unless stated otherwise.

Screening retains $K=10$ candidates, the enumeration considers subsets of size at
most $4$, and the invariance test uses $100$ permutations, or $200$ in the
population study of Section~\ref{sec:exp-verify}. Recovery experiments use eight
replicates; the calibration study of Table~\ref{tab:test-design} uses $150$
replicates for level and $100$ for power.

\section{Reproducing the results}
\label{app:repro}

The accompanying code contains the generator, the method, the baselines, and one
script per experiment reported in the paper. It requires Python~3.10 or later
with \texttt{numpy}, \texttt{scikit-learn}, \texttt{scipy} and
\texttt{matplotlib}; the results below were produced with Python~3.14,
\texttt{numpy}~2.2, \texttt{scikit-learn}~1.7, \texttt{scipy}~1.16 and
\texttt{matplotlib}~3.10.

Three modules carry the implementation. \texttt{scm.py} builds the
multi-environment structural causal models: a random DAG over $d$ covariates with
$s$ direct causes of $Y$ and three children, with sparse mechanism shifts per
environment, where \texttt{activate\_parents} selects the parent-activation
regime and \texttt{mech} selects the monotone or saturating mechanism.
\texttt{kirm.py} contains the method: \texttt{screen} is the $\ell_1$-MKL
screening stage, \texttt{invariance\_pvalue} the invariance test of
Section~\ref{sec:test-design}, and \texttt{kirm\_select} the enumeration
returning both selection rules; \texttt{invariance\_pvalue\_gcm} implements the
generalised covariance measure as one of the compared designs.
\texttt{baselines.py} contains pooled lasso, linear ICP, IRMv1, HSIC screening,
and the screening-only ablation.

Every experiment writes a result file, and two further scripts build the tables
and figures from those files. The result files used in the paper are included, so
\texttt{make\_tables.py} and \texttt{make\_figs.py} reproduce
Tables~\ref{tab:verify}, \ref{tab:nsweep}, \ref{tab:main} and~\ref{tab:diversity}
and Figures~\ref{fig:calibration} and~\ref{fig:recovery} without re-running
anything. The underlying results are regenerated one script at a time.
\texttt{ablate\_test.py} produces Table~\ref{tab:test-design}, which it writes
directly. \texttt{verify\_scaling.py} produces Table~\ref{tab:verify}.
\texttt{run\_experiments.py} takes an argument selecting the sweep, and
produces Table~\ref{tab:nsweep}, Table~\ref{tab:main} or
Table~\ref{tab:diversity} accordingly. \texttt{pool\_probe.py} produces the
experiment of Appendix~\ref{app:pool}, and \texttt{gcm\_ridge.py} the ridge
sensitivity quoted in Appendix~\ref{app:testdesign}. Each takes between ten
minutes and roughly an hour on a single core, the dimension sweep being the
longest, and all are single-threaded apart from the underlying BLAS calls.
Figure~\ref{fig:lattice} is drawn in the paper source and is not produced by the
code.

All scripts seed their generators explicitly and are deterministic for a fixed
\texttt{numpy} version. Small differences across \texttt{numpy} releases are
possible, since the random-number stream is version-dependent; this affects the
individual replicates rather than the rates reported in the tables.

Two properties of the generator are easy to lose when modifying it, and both were
failure modes we encountered. The edge probability must scale as $1/d$, for the
reason given in Appendix~\ref{app:generator}. And the environment interventions
must be applied to the covariates \emph{before} $Y$ is generated: drawing $Y$
first and shifting the parents afterwards leaves $P(Y \mid X_{\PA(Y)})$ varying
across environments, which violates Assumption~\ref{ass:modularity} and silently
removes the property the entire analysis depends on.

%% file: tables/main_table.tex
\begin{table*}[t]
\centering\small
\caption{Recovery as the ambient dimension grows ($n=2000$, five environments, $s=4$, eight replicates). `f.p.' is false positives per selection. Screening recall is $0.88$, $0.84$ and $0.75$ at $d=200$, $1000$ and $2000$.}
\label{tab:main}
\begin{tabular}{lccccccccc}
\toprule
 & \multicolumn{3}{c}{$d=200$} & \multicolumn{3}{c}{$d=1000$} & \multicolumn{3}{c}{$d=2000$} \\
\cmidrule(lr){2-4}\cmidrule(lr){5-7}\cmidrule(lr){8-10}
Method & Jac. & exact & f.p. & Jac. & exact & f.p. & Jac. & exact & f.p. \\
\midrule
Kernel ICP (minimal) & 0.37 & 0.25 & 1.62 & 0.50 & 0.38 & 0.62 & 0.14 & 0.00 & 1.62 \\
Kernel ICP (intersection) & 0.24 & 0.12 & 0.62 & 0.41 & 0.25 & 0.00 & 0.05 & 0.00 & 0.38 \\
MKL only (no invariance) & 0.63 & 0.25 & 1.00 & 0.62 & 0.12 & 1.00 & 0.61 & 0.25 & 1.12 \\
ICP (linear) & 0.62 & 0.62 & 0.00 & 0.50 & 0.50 & 0.00 & 0.38 & 0.38 & 0.00 \\
Lasso (pooled) & 0.40 & 0.00 & 6.00 & 0.40 & 0.00 & 6.00 & 0.40 & 0.00 & 6.00 \\
HSIC screening & 0.05 & 0.00 & 3.62 & 0.00 & 0.00 & 4.00 & 0.04 & 0.00 & 3.75 \\
IRMv1 (linear) & 0.00 & 0.00 & 4.00 & 0.00 & 0.00 & 4.00 & 0.00 & 0.00 & 4.00 \\
\bottomrule
\end{tabular}
\end{table*}

%% file: tables/diversity_table.tex
\begin{table}[t]
\centering\small
\caption{Jaccard index against the parent-activation regime ($d=1000$). `none' violates Assumption~\ref{ass:parent}, under which $\PA(Y)$ is genuinely not identifiable.}
\label{tab:diversity}
\begin{tabular}{lccc}
\toprule
Method & all & spread & none \\
\midrule
Kernel ICP (minimal) & 0.50 & 0.32 & 0.27 \\
Kernel ICP (intersection) & 0.41 & 0.15 & 0.03 \\
MKL only (no invariance) & 0.62 & 0.58 & 0.57 \\
ICP (linear) & 0.50 & 0.50 & 0.25 \\
Lasso (pooled) & 0.40 & 0.40 & 0.40 \\
HSIC screening & 0.00 & 0.02 & 0.05 \\
IRMv1 (linear) & 0.00 & 0.00 & 0.00 \\
\bottomrule
\end{tabular}
\end{table}